\documentclass[hidelinks, letterpaper, 10 pt, journal, twoside]{IEEEtran}

\ifCLASSINFOpdf
\else
\fi

\usepackage{graphicx}
\ifCLASSOPTIONcompsoc
\usepackage[caption=false,font=normalsize,labelfont=sf,textfont=sf]{subfig}
\else
\usepackage[caption=false,font=footnotesize]{subfig}
\fi
\usepackage{csquotes}

\usepackage[backend=bibtex,style=ieee, maxcitenames=2, mincitenames=1,natbib=true]{biblatex}
\usepackage{mathtools}
\usepackage{pifont}
\usepackage{gensymb}
\usepackage{makecell}
\newcommand{\cmark}{\ding{51}}%
\newcommand{\xmark}{\ding{53}}%
\usepackage[super]{nth}
\usepackage{textcomp}

\usepackage{hyperref}
\usepackage{amsfonts}
\usepackage{footnote}
\usepackage{booktabs}

\makesavenoteenv{tabular}
\begin{document}
%
\title{A Convex Optimization-based Dynamic Model Identification Package for the da Vinci Research Kit}

\author{Yan~Wang,~
        Radian~Gondokaryono,~
        Adnan~Munawar,~
        and~Gregory~S.~Fischer
\thanks{Authors are  with Automation and Interventional Medicine (AIM)
Lab, Worcester Polytechnic Institute, 100 Institute Rd, Worcester, MA, USA (Corresponding author: Gregory~S.~Fischer,  {\tt\small gfischer@wpi.edu})}
}

\maketitle
\begin{abstract}
The da Vinci Research Kit (dVRK) is a teleoperated surgical robotic system. 
For dynamic simulations and model-based control, the dynamic model of the dVRK is required. 
We present an open-source dynamic model identification package for the dVRK, capable of modeling the parallelograms, springs, counterweight, and tendon couplings, which are inherent to the dVRK. 
A convex optimization-based method is used to identify the dynamic parameters of the dVRK subject to physical consistency. 
Experimental results show the effectiveness of the modeling and the robustness of the package. 
Although this software package is originally developed for the dVRK, it is feasible to apply it on other similar robots. 
\end{abstract}


%
\IEEEpeerreviewmaketitle

\section{Introduction}
\IEEEPARstart{T}{he} da Vinci Research Kit (dVRK) is an open-source teleoperated surgical robotic system whose mechanical components are obtained from the first generation of the da Vinci Surgical Robot  \cite{kazanzides2014open}. 
It has made research on surgical robotics more accessible. 
To date, researchers from over 30 institutes\footnote{\url{http://research.intusurg.com/dvrkwiki}} around the world are using the physical dVRK, and some others are using the dVRK simulations \cite{kazanzides2014open, fontanelli2018v, richter2019open}.

Model-based control  has proven capable to increase the control precision and response speed of robotic arms \cite{reyes2001experimental}, as well as their capability to deal with surrounding environment \cite{de2006collision}.
Although these  techniques have already been widely used on traditional industrial robotic arms and collaborative robotic arms,
their research on surgical robots can be rarely found due to the lack of accurate dynamic models. 
Moreover,  several open-source simulators for the dVRK \cite{fontanelli2018v, richter2019open} have been  developed recently, which can potentially accelerate the development of robotic algorithms and surgical training.
However, accurate dynamic model, which is essential for realistic simulation, is absent in all of these simulators.

Several studies have been reported regarding the dynamic model identification of the dVRK \cite{fontanelli2017modelling, sang2017external, gondokaryono2018cooperative, pique2019dynamic}.
\citeauthor{fontanelli2017modelling} \cite{fontanelli2017modelling} identified the dynamic parameters of the Master Tool Manipulator (MTM) and Patient Side Manipulator (PSM) of the dVRK using the method proposed in \cite{sousa2014physical}.
In our previous work \cite{gondokaryono2018cooperative}, we replicated the approach in \cite{fontanelli2017modelling} and identified the dynamic parameters related to the first three joints of the PSM. 
With the obtained dynamic parameters, we implemented  a collaborative object manipulation based on impedance control, with one PSM manually controlled  by a user and the other following the former one's motion automatically.
\citeauthor{sang2017external} \cite{sang2017external} and \citeauthor{pique2019dynamic} \cite{pique2019dynamic} identified the dynamic parameters of the PSM using Least-square Regression and used them for sensorless external force estimation for force feedback.

Despite a significant amount of work regarding the dynamic model identification of the dVRK manipulators, none of them can be used directly by other researchers
since the dynamic parameters vary between different robots of the same make and model due to manufacturing and assembly variances.
Furthermore, the assembly components of robots are subject to deformation and wear \& tear along their life cycle, which can potentially alter the dynamic model.
As such, dynamic model identification is required before the implementation of any robust model-based control algorithm.
This requirement drives the need for a robust open-source dynamic model identification package.

There are existing software packages for the dynamic model identification of robotic manipulators, such as SymPybotics \cite{https://doi.org/10.5281/zenodo.11365},  FloBaRoID \cite{bethge2017flobaroid}, and OpenSYMORO \cite{khalil2014opensymoro}.
However, SymPybotics and  FloBaRoID are targeting at generic open-chain manipulators and lack the capability of modeling parallelograms, springs, counterweights, and tendon couplings, which are inherent to the mechanical design of the dVRK.
Although OpenSYMORO is able to model closed-chain mechanisms, no physical consistency (also called physical feasibility) \cite{wensing2018linear} is considered in parameter identification, 
which can potentially lead to unexpected behavior in simulations and model-based control \cite{yoshida2000verification}.
\begin{figure}
    \centering
    \includegraphics[width=0.29\textwidth]{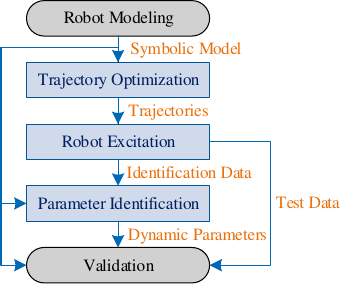}
    \caption{\label{fig:identification_workflow} Workflow of dynamic model identification.}
\end{figure}

The physical consistency conditions enforce the  positivity  of  kinetic  energy \cite{yoshida2000verification} and the  density  realizability of a link \cite{traversaro2016identification} by constraining the inertia tensor to be positive definite and the sum of any two of its eigenvalues to be larger than the third one.
These two conditions were formulated into semi-definite constraints with Linear  Matrix  Inequality  (LMI) techniques in \cite{sousa2014physical, sousa2019inertia, wensing2018linear}, enabling the use of convex optimization tools to solve the identification problem.
Moreover, \citeauthor{sousa2019inertia} \cite{sousa2019inertia} showed physical consistency constraints can  improve identification performance by reducing overfitting.

The purpose of this work is to develop an open-source dynamic model identification package for the dVRK considering full physical consistency.
Based on the workflow of dynamic model identification in Fig. \ref{fig:identification_workflow}, we structure this paper  into seven sections.
Sections \ref{sec_kinematics} and \ref{sec_dynamics} explain the mathematical formulation of the kinematic and dynamic modeling of the MTM and PSM.
Section \ref{sec:excitation_trajectory} describes the trajectory optimization method to improve parameter identification quality.
Section \ref{sec:identification} presents the identification approach to obtain physically consistent dynamic parameters.
The experimental results are presented to validate the proposed approaches in Section \ref{sec_result}.
The concluding arguments are entailed in Section \ref{sec:conclusion}.

\section{Kinematic Modeling of the dVRK}\label{sec_kinematics}

To build the relationship between the robot joint motion in the dVRK-ROS package \cite{kazanzides2014open} and the torque of each motor, several types of joint coordinates are defined. $\boldsymbol{q}^d$ are the joint coordinates used in the dVRK-ROS package. $\boldsymbol{q} = \begin{bmatrix}
(\boldsymbol{q}^b)^{\top} & \boldsymbol{(q}^a)^{\top}
\end{bmatrix}^{\top}$ are the joint coordinates used in the kinematic modeling in this work, where $\boldsymbol{q}^b$ are the basis joint coordinates which can adequately represent the kinematics of the robot, and $\boldsymbol{q}^a$ are the additional joint coordinates, which represent the other joint coordinates in the parallel mechanism and can be represented by the linear combination of $\boldsymbol{q}^b$. 
Since both the MTM and PSM have seven actuated degrees of freedom (DOF), the basis joint coordinates can be represented by $\boldsymbol{q}^b =
    \begin{bmatrix}
    q_1 & q_2 & \hdots & q_7
    \end{bmatrix}^\top$. $\boldsymbol{q}^{m}$ are the equivalent motor coordinates which are considered at joints, with the reduction ratio caused by gearboxes and tendons included for most motors unless explicitly specified. 
    Finally, 
$\boldsymbol{q}^{c} = \begin{bmatrix}
    \boldsymbol{q}^\top & (\boldsymbol{q}^m)^\top
    \end{bmatrix}^\top$ define the complete joint coordinates.
    The relation between these joint coordinates is illustrated for both the MTM and PSM in this section.
The dimensions  are  referred  from  the  user  guide  of  the  dVRK  or measured manually if not available.

\subsection{Kinematic Modeling of the MTM}

\begin{table}[!t]
    \caption{Modeling Description of the MTM }
    \label{table:mtm_geometry}
    \begin{center}
    \begin{tabular}{@{}l@{\hskip9pt}l@{\hskip9pt}l@{\hskip9pt}l@{\hskip9pt}l@{\hskip9pt}l@{\hskip9pt}l@{\hskip9pt}l@{\hskip9pt}l@{\hskip9pt}l@{}}
    \toprule
    $i$      & $a(i)$   & $a_{i-1}$ & $\alpha_{i-1}$     & $d_i$ & $\theta_i$ & $\boldsymbol{\delta}_{Li}$  & $I_{mi}$ & $\boldsymbol{F}_i$ & $K_{si}$\\\midrule
    1        & 0        & 0         & 0                  & $-l_{b2p}$ & $q_1$           & \cmark & \xmark & \cmark & \xmark\\
    2        & 1        & 0         & $-\frac{\pi}{2}$   & 0     & $q_2 + \frac{\pi}{2}$      & \cmark & \xmark & \cmark & \xmark\\
    3        & 2        & $l_{a}$   & 0                  & 0     & $q_3 + \frac{\pi}{2}$      & \cmark & \xmark & \cmark & \xmark\\
    $3'$     & 1        & 0         & $-\frac{\pi}{2}$   & 0     & $q_{3'} + \pi$      & \cmark & \xmark & \cmark & \xmark\\
    $3''$    & $3'$     & $l_{b2f}$ & 0                  & 0     & $-q_3 - \frac{\pi}{2}$    & \cmark & \xmark & \cmark & \xmark\\
    4        & 3        & $l_{f}$   & $-\frac{\pi}{2}$   & $h$ & $q_4$              & \cmark & \xmark & \cmark & \xmark\\
    5        & 4        & 0         & $\frac{\pi}{2}$    & 0     & $q_5$              & \cmark & \xmark & \cmark & \cmark\\
    6        & 5        & 0         & $-\frac{\pi}{2}$   & 0     & $q_6 + \frac{\pi}{2}$      & \cmark & \xmark & \cmark & \xmark\\
    7        & 6        & 0         & $-\frac{\pi}{2}$   & 0     & $q_7 + \pi$       & \cmark & \xmark & \cmark & \xmark\\
    $M_4$    & -        & 0         & 0                  & 0     & $q^d_4$          & \xmark & \cmark & \cmark & \xmark\\ \bottomrule
    \end{tabular}
    \end{center}
    \textbf{Note:} 
    $a(i)$ stands for the  antecedent link of link $i$. 
    $a_{i-1}$, $\alpha_{i-1}$, $d_i$, and $\theta_i$ are the modified DH parameters of link $i$.
    $\boldsymbol{\delta}_{Li}$, $I_{mi}$, $\boldsymbol{F}_i$, and $K_{si}$ are the parameters of link inertia, motor inertia, joint friction, and spring for link $i$, respectively.
    $M_4$ is an assistive frame for incorporating the joint coordinate of motor 4. 
    The other frames and used dimensions are shown in Fig. \ref{fig:mtm_frame_def}.

\end{table}

\begin{figure}[!t]
    \centering
    \includegraphics[width=0.27\textwidth]{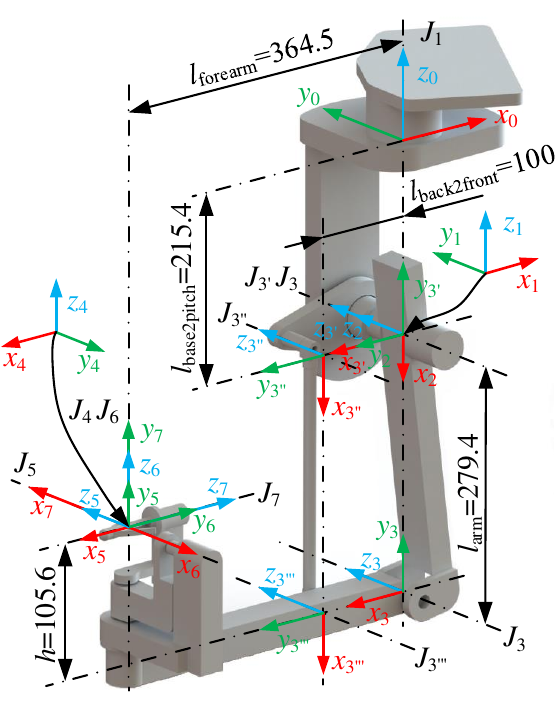}
    \caption{\label{fig:mtm_frame_def}Frame definition of the MTM using modified DH convention.}
\end{figure}

The left and right MTMs are identical to each other, except the last four joints being mirrored to each other. Consequently, the two MTMs can be modeled similarly.
The frame definition based on the modified Denavit-Hartenberg (DH) convention  \cite{khalil1986new}  is shown in Fig. \ref{fig:mtm_frame_def}, and the kinematic parameters of the MTM are described in Table \ref{table:mtm_geometry}. The kinematics of the MTM can be described as
\begin{itemize}
    \item Joint 1 rotates around the Z-axis of the base frame, $z_0$.
    \item Joints 2, 3, $3'$, $3''$, and $3'''$ construct a parallelogram, which is actuated by joints 2 and $3'$. 
    \item Joints 4, 5, 6, and 7 form a 4-axis non-locking gimbal.
\end{itemize}

The kinematics of the MTM is fully described by the basis joint coordinates $\boldsymbol{q}^b$, which are equal to the dVRK joint coordinate $\boldsymbol{q}^d$, $    \boldsymbol{q}^b = \boldsymbol{q}^d$.
The additional joints $\boldsymbol{q}^a$ can be described as the linear combination of $\boldsymbol{q}^b$ by
\begin{equation}\label{eq:mtm_coupling_model2closeloop}
\boldsymbol{q}^a = 
\begin{bmatrix}
q_{3'} &
q_{3''} &
q_{3'''}
\end{bmatrix}^\top =\begin{bmatrix}
q_2 + q_3 &
-q_3 &
q_3
\end{bmatrix}^\top
\end{equation}

\begin{figure}[!t]
\centering
\includegraphics[width=0.23\textwidth]{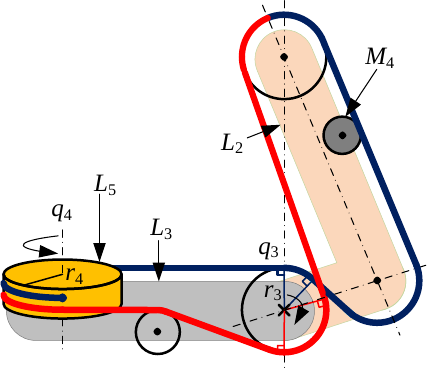}
\caption{\label{fig:mtm_tendon}Modeling of the tendon coupling of the MTM.}
\end{figure}

Joints 1, 5, 6, and 7 are independently driven, and thus the motion of these joints is equivalent to their corresponding driving motors, $\boldsymbol{q}^d_{1, 5-7}$ = $\boldsymbol{q}^m_{1, 5-7}$. The motion of $q^d_4$ depends on both $q^m_4$ and $q^d_3$ and can be described by 
\begin{equation}
q^d_4 = q^m_4 - {r_3}/{r_4}\cdot q^d_3
\end{equation}
where $r_3 \approx 14.01$ mm and $r_4 \approx 20.92$ mm are the radii of the pulleys shown in Fig. \ref{fig:mtm_tendon}.

The coupling between  $\boldsymbol{q}^d_{2-4}$ and $\boldsymbol{q}^m_{2-4}$ due to the parallelogram and tendons is resolved by the coupling matrix $\boldsymbol{A}^d_m$
\begin{equation}
    \boldsymbol{q}^d_{2-4} = \boldsymbol{A}^d_m \boldsymbol{q}^m_{2-4} = \small \begin{bmatrix}
    1 & 0 & 0\\ -1 & 1 & 0\\ 0.6697 & -0.6697 & 1 \end{bmatrix} \normalsize \boldsymbol{q}^m_{2-4}
\end{equation} 


\subsection{Kinematic Modeling of the PSM}

\begin{table}[!t]
    \caption{Modeling Description of the PSM
    }
    \label{table:psm_geometry}
    \begin{center}
    \begin{tabular}{@{}l@{\hskip9pt}l@{\hskip9pt}l@{\hskip9pt}l@{\hskip9pt}l@{\hskip9pt}l@{\hskip8pt}l@{\hskip8pt}l@{\hskip8pt}l@{\hskip8pt}l@{}}
    \toprule
    $i$       &  $a(i)$   & $a_{i-1}$ & $\alpha_{i-1}$ & $d_i$ & $\theta_i$  & $\boldsymbol{\delta}_{Li}$  & $I_{mi}$ & $\boldsymbol{F}_i$ & $K_{si}$\\\midrule
    1         & 0          & 0     & $\frac{\pi}{2}$    & 0 & $q_1+\frac{\pi}{2}$& \cmark & \xmark & \cmark & \xmark\\
    2         & 1       & 0     & $-\frac{\pi}{2}$   & 0     & $q_2 - \frac{\pi}{2}$& \cmark & \xmark & \cmark & \xmark\\
    $2'$      & 2         & $l_{2L3}$ & 0       & 0     & $\frac{\pi}{2}$& \xmark & \xmark & \xmark & \xmark\\
    $2''$     & $2'$        & $l_{2H1}$     & 0   & 0  & $\frac{\pi}{2} - q_2$ & \cmark & \xmark & \xmark & \xmark\\
    $2'''$    & $2'$        & $l_{c1}$    & 0  & 0     & $\frac{\pi}{2} - q_2$& \cmark & \xmark & \xmark & \xmark\\
    $2''''$   & $2''$         & $l_{2L2}$      & 0 & 0     & $q_2$& \cmark & \xmark & \xmark & \xmark\\
    $2'''''$  & $2''$        &  $l_{2L1}$    & 0   & 0     & $ q_2 + \pi$& \cmark & \xmark & \xmark & \xmark\\
    3         & $2''''$       & $l_3$      & $-\frac{\pi}{2}$  & $q_3 + l_{c2}$     & 0 & \cmark & \xmark & \cmark & \xmark\\
    $3'$      & 2       & $l_{2L3}$     & $-\frac{\pi}{2}$     & $q_3 $ & 0 & \cmark & \xmark & \xmark & \xmark\\
    4         & 3       &  0 & 0          & $l_{tool}$    & $q_4$& \xmark & \cmark & \cmark & \cmark\\
    5         & 4     &  0 & $\frac{\pi}{2}$          & 0     & $q_5 +\frac{\pi}{2}$ & \xmark & \cmark & \cmark & \xmark\\
    6         & 5      &  $l_{p2y}$ & $-\frac{\pi}{2}$    & 0     & $q_6+\frac{\pi}{2}$& \xmark & \xmark & \cmark & \xmark\\ 
    7         & 5       &  $l_{p2y}$ & $-\frac{\pi}{2}$         & 0     & $q_7+\frac{\pi}{2}$& \xmark & \xmark & \cmark & \xmark\\ 
    $M_6$     & -     &  0 & 0       & 0    & $q^m_6$ & \xmark & \cmark & \cmark & \xmark\\ 
    $M_7$     & -   &  0 & 0       & 0    & $q^m_7$ & \xmark & \cmark & \cmark & \xmark\\ 
    $F_{67}$  & -   &  0 & 0       & 0    & $q_6 - q_7$ & \xmark & \xmark & \cmark & \xmark\\ \bottomrule
    \end{tabular}
    \end{center}
    \textbf{Note:} Links 1 to 7 correspond to the links sescribed in Fig. 
    \ref{fig:psm_frame_def}. $M_6$ and $M_7$ correspond to the modeling of motors 6 and 7, respectively. 
    $F_{67}$ corresponds to the modeling of the relative motion between links 6 and 7. 
    The dimensions are shown in Fig. \ref{fig:psm_size_def}.
    $l_{c1} = l_{2H1}+l_{2H2}$, $l_{c2} = - l_{RCC} + l_{2H1}$.
\end{table}

\begin{figure}[!t]
    \centering
    \subfloat[\label{fig:psm_frame_def}Frame definition of the PSM using modified DH convention.]{\includegraphics[width=0.93\columnwidth]{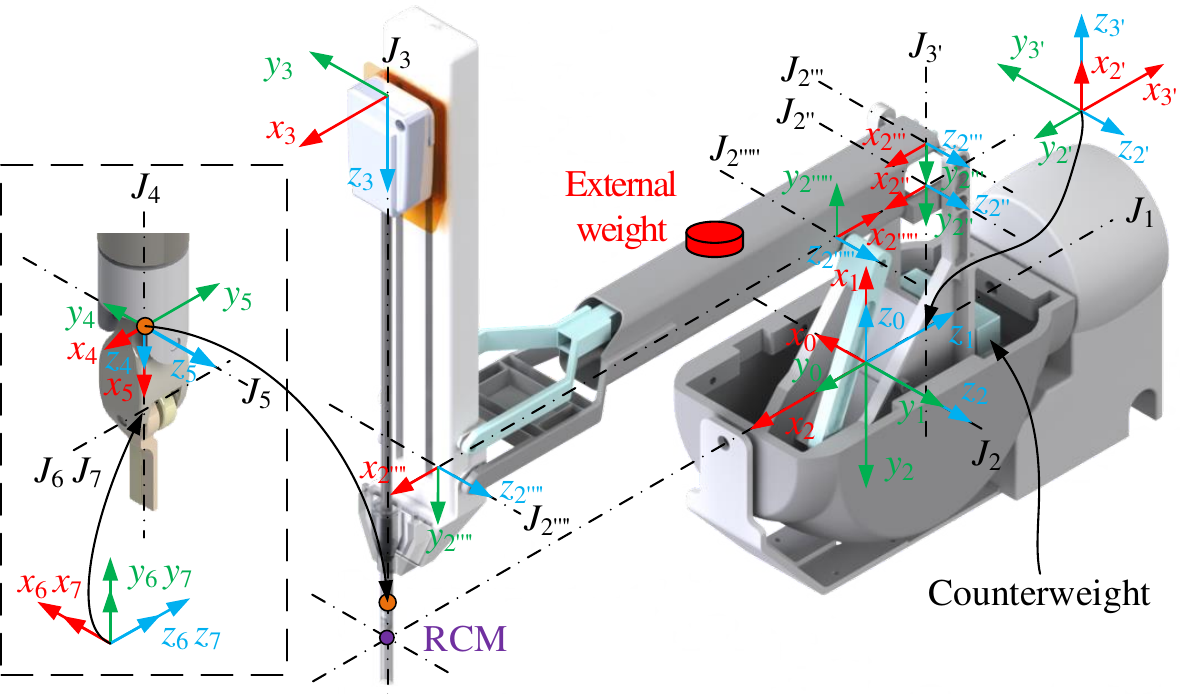}}\\
    
    \subfloat[\label{fig:psm_size_def}Planar view of the frame definition of the PSM.]{\includegraphics[width=0.92\columnwidth]{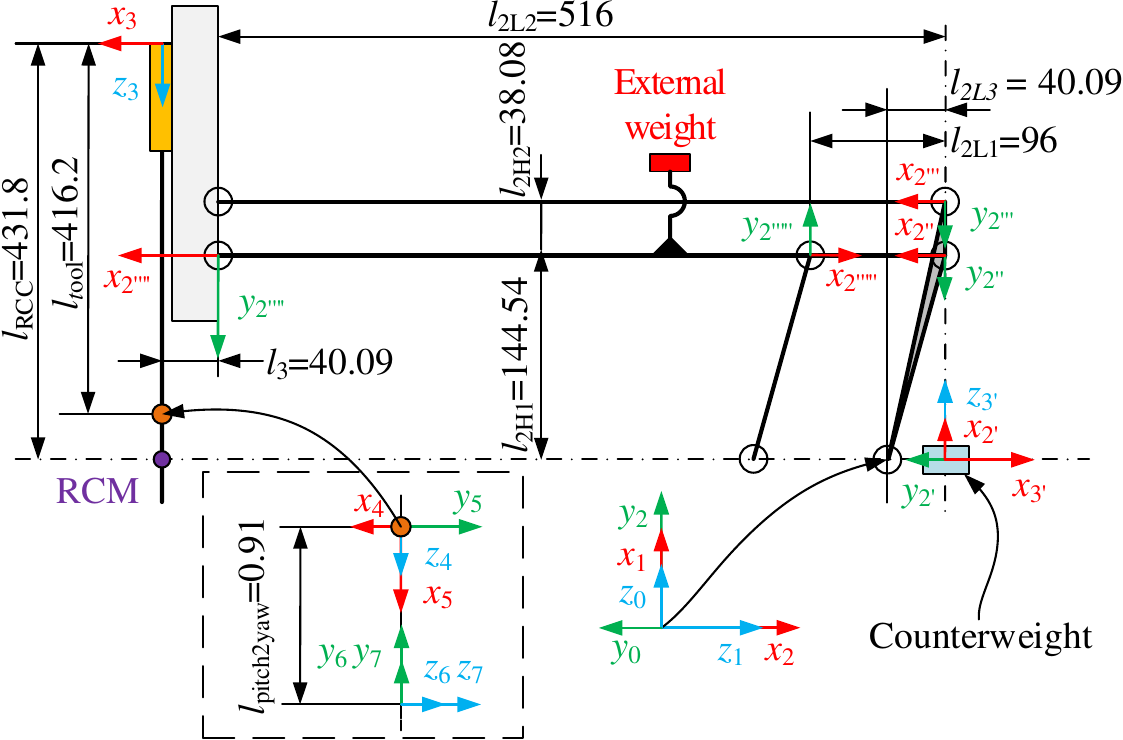}}

    \caption{\label{fig:psm_geo}
         Frame definition of the PSM.}
\end{figure}

The frame definition  of the PSM is shown in Fig. \ref{fig:psm_geo}, and the corresponding parameters are shown in Table \ref{table:psm_geometry}. The kinematics of the PSM can be concluded as

\begin{itemize}
    \item The first two revolute joints form a remote-center-of-motion (RCM) point  via a double four-bar linkage with six links actuated by a single motor. 
    \item The third joint is prismatic and provides the insertion of the instrument through the RCM. The first three joints allow the 3-DOF Cartesian space motion.
    \item Revolute joints 4 and 5 construct the roll and pitch motion of the wrist to reorient the end-effector. 
    \item The last two joints construct the yaw motion of the end-effector, as well as the opening and closing of the gripper. 
\end{itemize}

\begin{figure}[!tb]
    \centering
            \subfloat[\label{fig:psm_gripper_motion_modeling}Modeling of the motion of the gripper.]{\includegraphics[width=0.34\columnwidth]{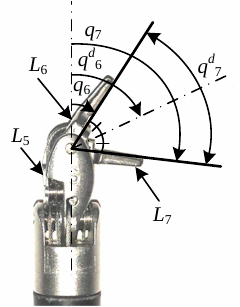}}
        \qquad
            \subfloat[\label{fig:psm_gripper_friction_modeling}Modeling of the frictions of the gripper.]{\includegraphics[width=0.32\columnwidth]{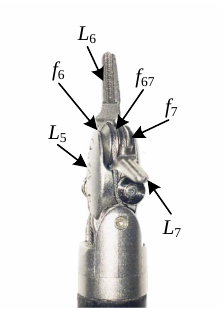}}
    
    \caption{\label{fig:psm_gripper}Modeling of the gripper of the PSM.}
\end{figure}

We model the first five joints of the PSM identical to the dVRK-ROS package, i.e., $\boldsymbol{q}_{1-5} = \boldsymbol{q}^d_{1-5}$. 
The dVRK-ROS package models the last two joints as $q^d_6$, the angle from the insertion axis to the bisector of the two jaw tips, and $q^d_7$, the angle between the two jaw tips. 
However, the gripper jaws are designed and actuated as two separate links. 
As shown in Fig. \ref{fig:psm_gripper_motion_modeling}, the relation between  $\boldsymbol{q}^d_{6-7}$, and  $\boldsymbol{q}_{6-7}$  is described by
\begin{equation}\label{eq:psm_coupling_model2dvrk}
\boldsymbol{q}^d_{6-7} = 
\begin{bmatrix}
q^d_{6} &
q^d_{7}
\end{bmatrix}^\top =\begin{bmatrix}
0.5q_{6} + 0.5q_{7} &
-q_{6}+q_{7}
\end{bmatrix}^\top
\end{equation}

Since the first four joints are independently driven, the equivalent motor motion is considered to occur at joints,
i.e., $\boldsymbol{q}^d_{1-4} = \boldsymbol{q}^m_{1-4}$. 
Based on the user guide of the dVRK, the coupling of the wrist joint actuation can be resolved by the coupling matrix $\boldsymbol{A}^d_m$ mapping $\boldsymbol{q}^m_{5-7}$ to  $\boldsymbol{q}^d_{5-7}$ by
\begin{equation}\label{eq:psm_coupling}
\boldsymbol{q}^d_{5-7} = \boldsymbol{A}^d_m \boldsymbol{q}^m_{5-7}
=\small \begin{bmatrix}
    1.0186 & 0 & 0\\
    -0.8306 & 0.6089 & 0.6089\\
    0 & -1.2177 & 1.2177
    \end{bmatrix} \normalsize \boldsymbol{q}^m_{5-7}
\end{equation}

\section{Dynamic Modeling of the dVRK}\label{sec_dynamics}
In this section, the dynamic parameters are described first. The dynamic equation is then formulated based on the Euler-Lagrange equation. Finally, the dynamic modeling of the MTM and PSM is introduced based on the formulation.

\subsection{Dynamic Parameters}

Each link $k$ is characterized by the mass $m_k$, the center of mass (COM) relative to the link frame $k$, $
\boldsymbol{r}_k$,
and the inertia tensor about the COM, $\boldsymbol{I}_k$.
To express the equations of motion as a linear form of dynamic parameters, we use the so-called barycentric parameters \cite{maes1989linearity}, in which the mass $m_k$ of link $k$ is first used, followed by the first moment of inertia, $\boldsymbol{l}_k =m_k \boldsymbol{r}_k$.
Finally, the inertia tensor $\boldsymbol{L}_k$ about frame $k$ is used \cite{khalil2004modeling}, which is calculated via the parallel axis theorem
\begin{equation}\label{eq:Iml2L}
\boldsymbol{L}_k = \boldsymbol{I}_k + m_k \boldsymbol{S}(\frac{\boldsymbol{l}_k}{m_k})^\top\boldsymbol{S}(\frac{\boldsymbol{l}_k}{m_k}) = 
\small
\begin{bmatrix}
L_{kxx} & L_{kxy} & L_{kxz}\\
L_{kxy} & L_{kyy} & L_{kyz}\\
L_{kxz} & L_{kyz} & L_{kzz}
\end{bmatrix} 
\normalsize
\end{equation}
where $\boldsymbol{S}(\cdot)$ is the skew-symmetric operator.

The aforementioned  inertial parameters of link $k$ are grouped into a vector $\boldsymbol{\delta}_{Lk} \in \mathbb{R}^{10}$ as
\begin{equation}
\label{eq:inertia}
\boldsymbol{\delta}_{Lk} = \small [\begin{matrix} L_{kxx} & L_{kxy} & L_{kxz} & L_{kyy} & L_{kyz} & L_{kzz} &
 \boldsymbol{l}_{k}^{\top} & m_k \end{matrix}]^\top \normalsize
\end{equation}

Besides the inertial parameters of link $k$, the corresponding joint friction coefficients, motor inertia $I_{mk}$, and spring stiffness $K_{sk}$ are grouped as additional parameters
\begin{equation}
\boldsymbol{\delta}_{Ak} = \begin{bmatrix}
F_{vk} & F_{ck} & F_{ok} & I_{mk} & K_{sk}
\end{bmatrix}^\top
\end{equation}
where $F_{vk}$ and $F_{ck}$ are the viscous and Coulomb friction constants, and $F_{ok}$ is the Coulomb friction offset of joint $k$.


Eventually, all the parameters of $n$ joints are grouped together as the dynamic parameters $\boldsymbol{\delta}$ of a robot.
\begin{equation}
\label{eq:parameters}
\boldsymbol{\delta} = 
\begin{bmatrix}
\boldsymbol{\delta}_{L1}^\top & \boldsymbol{\delta}_{A1}^\top & ... & \boldsymbol{\delta}_{Ln}^\top & \boldsymbol{\delta}_{An}^\top
\end{bmatrix}^{\top}
\end{equation}

\subsection{Dynamic Model Formulation}
The inverse dynamic model for closed-chain robots, which relates motor torques and joint motion, can be calculated using Newton-Euler \cite{khalil2010dynamic} or Euler-Lagrange \cite{nakamura1989dynamics} methods for the equivalent tree structure and by considering kinematic constraints between joint coordinates.
The Euler-Lagrange equation is used to model the dynamics of the dVRK, due to its ease of dealing with kinematic constraints. 
The Lagrangian is calculated by the difference of the kinetic energy  $K$  and potential energy $P$ of the robot, $L = K - P$. 
Motor inertias and springs are not included in $L$ and modeled separately. 

The relation from motor motion $\boldsymbol{q}^m$ to the torque  of each motor $i$ caused by link inertia is then computed as
\begin{equation}
\tau^m_{LIi} = \frac{\mathrm{d}}{\mathrm{d}t} \frac{\partial L}{\partial \dot{q}^m_{i}} - \frac{\partial L}{\partial q^m_{i}}
\end{equation}

The friction torques of all the joints $\boldsymbol{q}^{c}$ are considered as 
\begin{equation} \label{eq:friction}
\boldsymbol{\tau}^{c}_f(\dot{\boldsymbol{q}}^{c}) = \boldsymbol{F}_v \dot{\boldsymbol{q}}^{c} + \boldsymbol{F}_c \boldsymbol{\mathrm{sgn}}(\dot{\boldsymbol{q}}^{c}) + \boldsymbol{F}_o
\end{equation}
where $\boldsymbol{F}_v$ and $\boldsymbol{F}_c$ are diagonal matrices encapsulating the viscous and Coulomb friction constants, and $\boldsymbol{F}_o$ is the vector of the Coulomb friction offset constants.

The torques caused by motor inertia are defined as 
\begin{equation}
\boldsymbol{\tau}^m_{MI}(\ddot{\boldsymbol{q}}^m) = \boldsymbol{I}_m \ddot{\boldsymbol{q}}^m
\end{equation}

For spring $k$, we only model the stiffness constant $K_{sk}$ as its parameter, which results into the spring torques 
\begin{equation}
\boldsymbol{\tau}^c_s(\boldsymbol{q}^{c}) = \boldsymbol{K}_s {\boldsymbol{\Delta l}_s}
\end{equation}
where $\boldsymbol{K}_s$ is a diagonal matrix of the stiffness constants of the springs, and $\boldsymbol{\Delta l}_s$ is the equivalent prolongation vector.

The joint torques caused by springs and frictions can be projected onto the motor joints, using the Jacobian matrix of their corresponding joint coordinate with respect to the motor joint angle $\boldsymbol{q}^m$ \cite{nakamura1989dynamics}. Thus, the motor torques $\boldsymbol{\tau}^m$ with link inertia, springs, frictions, motor inertia, and motion couplings considered are given by 
\begin{equation} \label{eq:close_chain_torque}
\boldsymbol{\tau}^m = \boldsymbol{\tau}^m_{LI}  + \boldsymbol{\tau}^m_{MI}(\ddot{\boldsymbol{q}}^m)  + \frac{\partial{\boldsymbol{q}^{c}}}{\partial\boldsymbol{q}_m}(\boldsymbol{\tau}^c_s(\boldsymbol{q}^{c}) + \boldsymbol{\tau}^{c}_f(\dot{\boldsymbol{q}}^{c}))
\end{equation}

To identify $\boldsymbol{\delta}$, \eqref{eq:close_chain_torque} is rewritten into \eqref{eq:close_chain_torqe_linear} by the linear parameterization.
\begin{equation} \label{eq:close_chain_torqe_linear}
\boldsymbol{\tau}^m = \boldsymbol{H}(\boldsymbol{q}^m, \dot{\boldsymbol{q}}^m, \ddot{\boldsymbol{q}}^m) \boldsymbol{\delta}
\end{equation}


QR decomposition with pivoting \cite{gautier1991numerical} is used to calculate the base parameters, a minimum set of dynamic parameters that can fully describe the dynamic model of a robot. 
With this method, we get a permutation matrix $\boldsymbol{P}_\mathrm{b} \in \boldsymbol{\mathbb{R}}^{n\times b}$, where $n$ is the number of standard dynamic parameters and $b$ is the number of base parameters. 
The base parameters $\boldsymbol{\delta}_{\mathrm{b}}$ and the corresponding regressor $\boldsymbol{H}_\mathrm{b}$ can then be calculated by 
\begin{equation} \label{eq:base_parameters}
    \boldsymbol{\delta}_{\mathrm{b}} = \boldsymbol{P}_\mathrm{b}^{\top}\boldsymbol{\delta},\,\,\,\,  \boldsymbol{H}_{\mathrm{b}} = \boldsymbol{H}\boldsymbol{P}_\mathrm{b}
\end{equation}

\subsection{Dynamic Modeling of the MTM}
\begin{figure}[!t]
    \centering
            \subfloat[\label{fig:mtm_cable}Electrical cable on joint 4 of the MTM.]{\includegraphics[width=0.35\columnwidth]{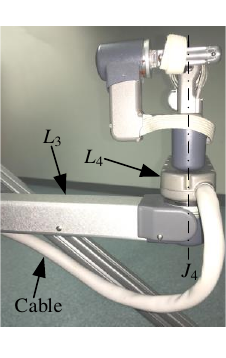}}
        \qquad
            \subfloat[\label{fig:mtm_cable_modeling}Modeling of the joint torque from the electrical cable on joint 4.]{\includegraphics[width=0.46\columnwidth]{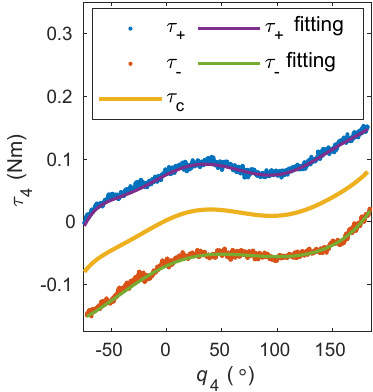}}
    
    \caption{Modeling of the joint torque from the electrical cable on joint 4 of the MTM.}
\end{figure}

The dynamic modeling description for  the MTM is shown in Table \ref{table:mtm_geometry}. 
All the nine links are modeled with link inertia. 
The frictions of all the joints $\boldsymbol{q}$ are considered, except joint $3'''$ since joint $3'''$ and joint $3''$ share the same joint coordinate, and their frictions are coupled together. 
Similarly, all the motors except the $\nth{4}$ one have their corresponding independently driven joints which have already been modeled with link inertia and joint friction.  
Therefore, only motor 4  is modeled with motor inertia and motor friction. 

The electrical cable along joint 4 (Fig. \ref{fig:mtm_cable}) affects its joint torque significantly. The joint torque data of joint 4, $\tau_{4}^+$ and $\tau^-_4$, was collected, with joint 4 rotating at $\pm 0.4$ rad/s and other joints being stationary, as shown in Fig. \ref{fig:mtm_cable_modeling}. 
We collected data at constant joint velocities, which explicitly removes any torque due to inertia. 
Moreover, due to the friction model in \eqref{eq:friction}, the frictions with the joint velocity at $\pm 0.4$ rad/s should be opposite to each other if the Coulomb friction offset is not considered. 
Thus finally, we computed the mean of $\tau_{4}^+$ and $\tau^-_4$, which canceled the viscous and Coulomb friction terms and kept the joint friction offset and torque applied to the joint from the cable physically acting on it, $\tau^m_{c4}(q_4)$. 

To get $\tau^m_{c4}(q_4)$, we first fitted the joint torque data at $\pm 0.4$ rad/s using $\nth{7}$ order polynomial functions of $q_4$, respectively, as shown in Fig. \ref{fig:mtm_cable_modeling}. Next, the mean of the obtained coefficients of the two polynomials $\boldsymbol{p}_4^+$ and $\boldsymbol{p}_4^-$ was calculated as the coefficients of the polynomial that represents $\tau^m_{c4}(q_4)$.


\begin{figure}[!t]
    \centering
    
        \subfloat[\label{fig:mtm_right_spring_picture}Spring on joint 5.]{\includegraphics[width=0.33\columnwidth]{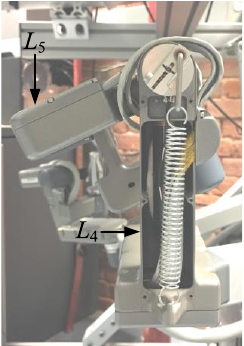}}
        \qquad
        \subfloat[\label{fig:mtm_right_spring_modeling}Modeling of the spring.]{\includegraphics[width=0.35\columnwidth]{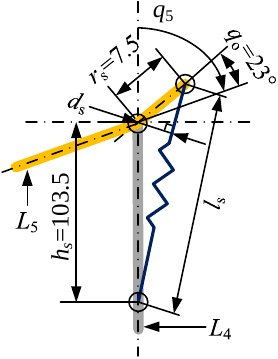}}

    \caption{\label{fig:mtm_right_spring}Spring on joint 5 of the MTM and its modeling.}
\end{figure}

In addition, on joint 5 of the MTM, there is a spring to balance the gravitational force (Fig. \ref{fig:mtm_right_spring_picture}).
Due to the modeling shown in Fig. \ref{fig:mtm_right_spring_modeling}, the joint torque from the spring is given by
\begin{equation}
\tau_{s5} = f_{s} \cdot d_{s} = K_{s5}(l_s - l_r) \cdot d_{s} = K_{s5}\Delta l_{s5}
\end{equation}
where $l_s$ is the length between the two axes connecting the spring, which can be calculated using the law of sines as
\begin{equation}
l_s = \sqrt[]{h_s^2 + r_s^2 - 2 h_s r_s \cos(\pi + q_o - q_5)}
\end{equation}
and $l_r \approx 61.3$ mm by measurement is the value of $l_s$ when the spring is relaxed.
Based on basic trigonometry, the moment arm $d_{s}$ can be calculated by 
\begin{equation}
d_{s} = h_s r_s \sin(\pi + q_o - q_5) / l_s
\end{equation}
where $h_s$, $r_s$ and $q_o$ are constants shown in Fig. \ref{fig:mtm_right_spring_modeling}.

Thus, $\Delta l_{s5} = (l_s - l_r)d_{s}$.

\subsection{Dynamic Modeling of the PSM}
The dynamic modeling description of the PSM is shown in Table \ref{table:psm_geometry}. Inertia is considered for all the links contributing to the Cartesian motion, including the counterweight, link $3'$. The motor inertia of these joints is ignored since it is not significant compared to their link inertia. The inertia of the wrist and gripper links is minimal, and thus infeasible to identify. Therefore, we only model the inertia of motors for the wrist and gripper, corresponding to the motion of $\boldsymbol{q}^m_{4-7}$.

Since joints $2$, $2''$, $2'''$, $2''''$, and $2'''''$ are all driven by a single motor, their frictions can be represented by the friction of one joint for simplicity. 
Thus, among these joints, only joint 2 is modeled with friction. 
Similarly, only joint 3 is modeled with friction out of joints 3 and $3'$. 
Because of the contact between links 5 and 6, and between links 5 and 7, as shown in Fig. \ref{fig:psm_gripper_friction_modeling}, the frictions on joints 6 and 7 are modeled, corresponding to the motion of $q_6$ and $q_7$. 
Moreover, the friction between links 6 and 7 due to the contact between the two jaw tips is considered, corresponding to the motion of $q_7-q_6$. 
Additionally, the frictions on the motor sides of the last four joints are also modeled, corresponding to the motor motion of $\boldsymbol{q}^m_{4-7}$.

The torsional spring on joint 4 which rotates the joint back to its home position is modeled as
\begin{equation}
\tau_{s4} = K_{s4}(-q_4) = K_{s4}\Delta l_{s4}
\end{equation}

\section{Excitation Trajectory Optimization}\label{sec:excitation_trajectory}
Periodic excitation trajectories based on Fourier series    \cite{swevers1997optimal} are used to generate  data for dynamic model identification. 
These trajectories minimize the condition number of the regression matrix $\boldsymbol{W}_b$ for the base parameters $\boldsymbol{\delta}_\mathrm{b}$, which decide the dynamic behavior of a robot.
\begin{equation}
\label{eq:reg_base}
    \boldsymbol{W}_b =
\small
 \begin{bmatrix}
\boldsymbol{H}_b(\boldsymbol{q}^m_1, \dot{\boldsymbol{q}}^m_1, \ddot{\boldsymbol{q}}^m_1)\\
\boldsymbol{H}_b(\boldsymbol{q}^m_2, \dot{\boldsymbol{q}}^m_2, \ddot{\boldsymbol{q}}^m_2)\\
\vdots\\
\boldsymbol{H}_b(\boldsymbol{q}^m_{S}, \dot{\boldsymbol{q}}^m_{S}, \ddot{\boldsymbol{q}}^m_S)\\
\end{bmatrix}
\normalsize
\end{equation}
where $\boldsymbol{q}^m_{i}$ is the motor joint coordinate at $i^{\mathrm{th}}$ sampling point and $S$ is the number of sampling points. 

The joint coordinate $q^m_{k}$ of motor $k$ can be calculated by
\begin{equation}
q^m_{k}(t) = q^m_{ok} + \sum_{l=1}^{n_\mathrm{H}} \frac{a_{lk}}{\omega_f l} \sin(\omega_f l t) - \frac{b_{lk}}{\omega_f l}\cos(\omega_f l t)
\end{equation}
where $\omega_f = 2\pi f_f$ is the angular component of the fundamental frequency $f_f$, $n_\mathrm{H}$ is the harmonic number of Fourier series, $a_{lk}$ and $b_{lk}$ are the amplitudes of the $l^{\mathrm{th}}$-order sine and cosine functions, $q^m_{ok}$ is the position offset, and $t$ is the time.

The motor joint velocity $\dot{q}^m_k(t)$ and acceleration $\ddot{q}^m_k(t)$ can be calculated easily by the differentiation of $q^m_k(t)$. And the trajectory must satisfy the following constraints:
\begin{itemize}
\item The joint position $\boldsymbol{q}$ is between the lower bound $\boldsymbol{q}_l$ and the upper bound $\boldsymbol{q}_u$, $\boldsymbol{q}_l \le \boldsymbol{q} \le \boldsymbol{q}_u$.
\item The absolute value of the joint velocity $\dot{\boldsymbol{q}}$ is smaller than its maximum value $\dot{\boldsymbol{q}}_{max}$, $|\dot{\boldsymbol{q}}| < \dot{\boldsymbol{q}}_{max}$.
\item The robot is confined in its workspace. The Cartesian position $\boldsymbol{p}_k$ of frame $k$ is between its lower bound $\boldsymbol{p}_{lk}$ and upper bound $\boldsymbol{p}_{uk}$, $\boldsymbol{p}_{lk} \le \boldsymbol{p}_k \le \boldsymbol{p}_{uk}$.
\end{itemize}


\section{Identification}\label{sec:identification}
To identify the dynamic parameters, we move the robot along the excitation trajectories generated via the method described in Section \ref{sec:excitation_trajectory}. 
Data is collected at each sampling time to obtain the regression matrix $\boldsymbol{W}$ and the dependent variable vector $\boldsymbol{\omega}$.
\begin{equation}
\label{eq:reg}
    \boldsymbol{W} =
\small \begin{bmatrix}
\boldsymbol{H}(\boldsymbol{q}^m_1, \dot{\boldsymbol{q}}^m_1, \ddot{\boldsymbol{q}}^m_1)\\
\boldsymbol{H}(\boldsymbol{q}^m_2, \dot{\boldsymbol{q}}^m_2, \ddot{\boldsymbol{q}}^m_2)\\
\vdots\\
\boldsymbol{H}(\boldsymbol{q}^m_\mathrm{S}, \dot{\boldsymbol{q}}^m_S, \ddot{\boldsymbol{q}}^m_S)\\
\end{bmatrix} \normalsize,\:
\boldsymbol{\omega} =
 \small \begin{bmatrix}
\boldsymbol{\tau}^m_{1} \\
\boldsymbol{\tau}^m_{2} \\
\vdots \\
\boldsymbol{\tau}^m_{S}\\
\end{bmatrix}
\normalsize
\end{equation}
where $\boldsymbol{\tau}^m_{i}$ is the motor torque at $i^{\mathrm{th}}$ sampling point.

The identification problem can then be formulated into an optimization problem which minimizes the squared residual error $||\boldsymbol{\epsilon}||^2$ w.r.t. the decision vector $\boldsymbol{\delta}$.
\begin{equation} \label{eq:residual_error}
||\boldsymbol{\epsilon}||^2 = ||\boldsymbol{W} \boldsymbol{\delta} - \boldsymbol{\omega}||^2
\end{equation} 

To get more realistic dynamic parameters and reduce overfitting problems \cite{sousa2019inertia}, we utilize physical consistency constraints  for dynamic parameters:
\begin{itemize}
    \item The mass of each link $k$ is positive, $m_k > 0$.
    \item The inertia matrix of each link $k$ is positive definite, $\boldsymbol{I}_k \succ \boldsymbol{0}$ \cite{yoshida2000verification}, and its eigenvalues, $Y_x$, $Y_y$, and $Y_z$, should follow the so-called triangle inequality conditions \cite{traversaro2016identification}, i.e., $Y_x + Y_y > Y_z$, $Y_y + Y_z > Y_x$, and $Y_z + Y_x > Y_y$.
    \item The COM of link $k$, $\boldsymbol{r}_k$, is inside its convex hull, $m_k\boldsymbol{r}_{lk} - \boldsymbol{l}_k \le 0$ and $m_k\boldsymbol{r}_{uk} + \boldsymbol{l}_k \le 0$, where $\boldsymbol{r}_{lk}$ and $\boldsymbol{r}_{uk}$ are the lower and upper bounds of $\boldsymbol{r}_{k}$, respectively \cite{sousa2014physical}.
    \item The viscous and Coulomb friction coefficients for each joint $i$ are positive, $F_{vi} > 0$ and $F_{ci} > 0$.
    \item The inertia of motor $k$  is positive, $I_{mk} > 0$.
    \item The stiffness of spring $k$ is positive, $K_k > 0$.
\end{itemize}

The first two constraints regarding the inertia properties of link $k$ can be derived into an equivalent  with LMIs \cite{wensing2018linear} as
\begin{equation}
  \boldsymbol{\bar{D}}_k(\boldsymbol{\delta}_{Lk}) = \small \begin{bmatrix}
\frac{1}{2}\mathrm{tr}(\boldsymbol{L}_{k})\cdot \boldsymbol{1}_3 - \boldsymbol{L}_{k}  & \boldsymbol{l}_k\\
\boldsymbol{l}_k^{\top} & m_k
\end{bmatrix} \normalsize \succ \boldsymbol{0}  
\end{equation}

We can also add the lower and upper bounds to $m_k$, $F_{vi}$, $F_{ci}$ and $K_j$ when we have more knowledge about them.

\section{Experiments}\label{sec_result}
This section presents the experimental procedures and results of the dynamic model identification of the dVRK arms.
\subsection{Experimental Procedures}
\subsubsection{Excitation Trajectory Generation}
Two independent excitation trajectories were generated for identification and test, respectively, for each of the MTM and PSM. The harmonic number $n_H$ was set to 6.
The fundamental frequency $f_f$ of the MTM and PSM were 0.1 and 0.18 Hz, respectively. 
The joint position and velocity were constrained within their ranges in the optimization.
Since links $2''$ and $2'''$ of the PSM  are very close to each other with similar motion, it is hard to get a trajectory with a low condition number of $\boldsymbol{W}_b$ when both links $2''$ and $2'''$ are considered. 
Links 2 and $2'''''$ have the similar problem. 
Therefore, the trajectory optimization of the PSM was based on the model without links $2'''$ and $2'''''$. 
Finally, pyOpt \cite{perez2012pyopt} was used to solve this constrained nonlinear optimization problem.

\subsubsection{Data Collection and Processing}
 The joint position, velocity, and torque were collected at 200 Hz in position control mode.
The joint acceleration was obtained by the second-order numerical differentiation of the velocity. 
A sixth-order low-pass filter was used to filter the data with the cutoff frequencies of 1.8 Hz for the MTM and 5.4 Hz for the PSM. 
The cutoff frequencies were chosen experimentally to achieve the best identification performance as they are low enough to filter the noise as well as high enough to keep the useful signal in the collected data. 
To achieve zero phase delay, we applied this filter in both forward and backward directions.


    



\subsubsection{Identification}
To get uniformly precise identification results for all joints, the residual error $\boldsymbol{\epsilon}_i$ of each motor joint $i$ in \eqref{eq:residual_error} was weighted by $w_i = 1/(\max\{\boldsymbol{\tau}^m_i\} - \min\{\boldsymbol{\tau}^m_{i}\})$. 
As a convex optimization problem, the identification was solved via the CVXPY package \cite{cvxpy} with the SCS solver \cite{scs}.
\subsection{Validation of the Identified Values}
\subsubsection{Identification for the dVRK Arms}
The identified dynamic parameters $\hat{\boldsymbol{\delta}}$ from identification trajectories were used to predict the motor joint torque on  test trajectories, $\hat{\boldsymbol{\omega}} = \boldsymbol{W}\hat{\boldsymbol{\delta}}$. 
The relative root mean squared error was used as the relative prediction error to assess the identification quality,
$\boldsymbol{\epsilon} = ||\boldsymbol{\omega} - \hat{\boldsymbol{\omega}}||_2/{||\boldsymbol{\omega}||_2}$.
The same experimental procedure was conducted with the modeling from \cite{fontanelli2017modelling} for comparision since it is the only previous work considering physical consistency.

\begin{table}[!tb]
\caption{Relative Prediction Error on Test Trajectories}
\label{table:relative_predict_error}
\begin{center}
\begin{tabular}{@{}l@{\hskip12pt}l@{\hskip12pt}l@{\hskip12pt}l@{\hskip12pt}l@{\hskip12pt}l@{\hskip12pt}l@{\hskip12pt}l@{}}
\toprule
 & $\tau^m_1$ & $\tau^m_2$ & $\tau^m_3$ or $f^m_3$ & $\tau^m_4$ & $\tau^m_5$ & $\tau^m_6$ & $\tau^m_7$\\
\midrule
MTM-Y (\%) & 7.6     & 14.9   &  17.0 & 22.3 & 28.0 & 23.4 & 34.0 \\
MTM-F (\%) & 11.5     & 18.6   &  40.0 & 36.2 & 69.3 & 31.1 & 37.0 \\
PSM-Y (\%)  & 9.3 & 17.8   & 19.1 & 13.4 & 23.9 & 21.3 & 26.4\\
PSM-F (\%)  & 10.6 & 18.8   & 18.9 & 88.7 & 87.8 & 72.2 & 36.5\\
\bottomrule
\end{tabular}
\end{center}
\end{table}
\begin{figure}[!ht]
\hfil
\hfil
\includegraphics[width=0.48\textwidth]{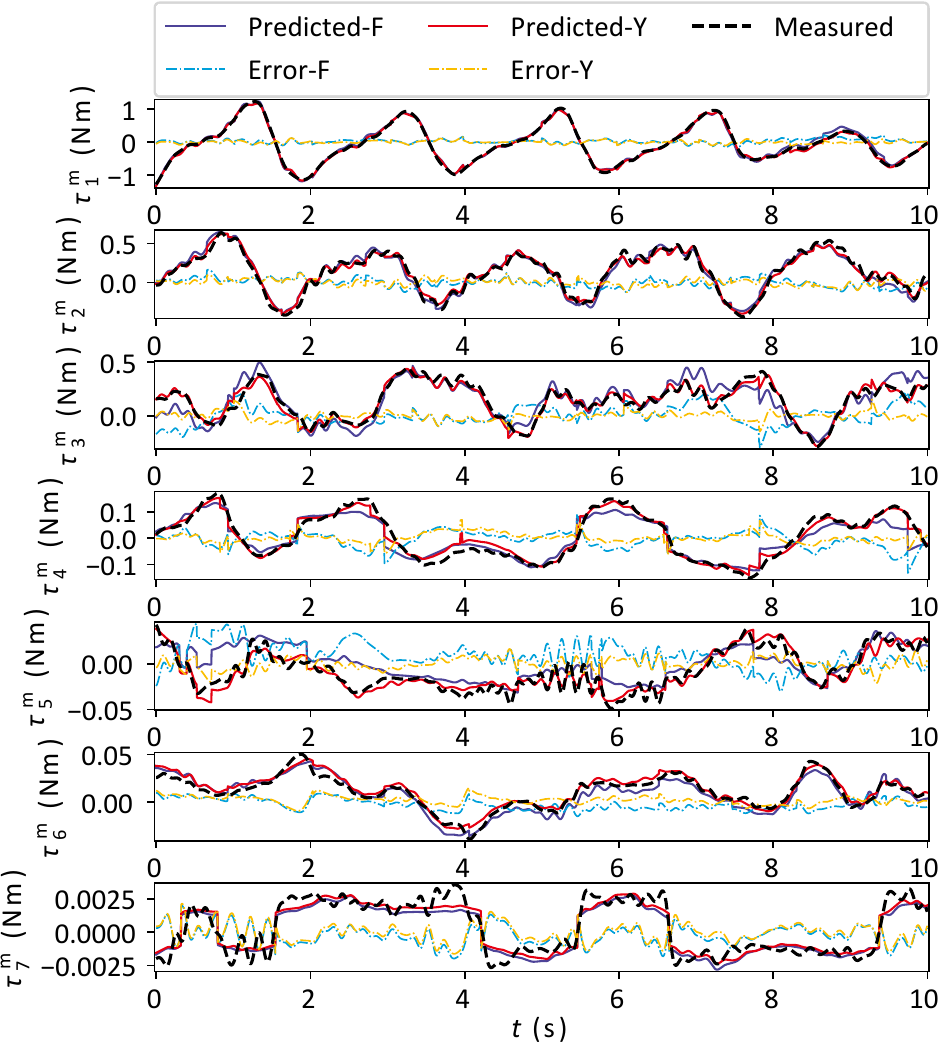}
\caption{\label{fig:mtm_cvx_test} Measured and  predicted torques on the test trajectory of the MTM.  }
\end{figure}
\begin{figure}[!ht]
\hfil
\hfil
\includegraphics[width=0.48\textwidth]{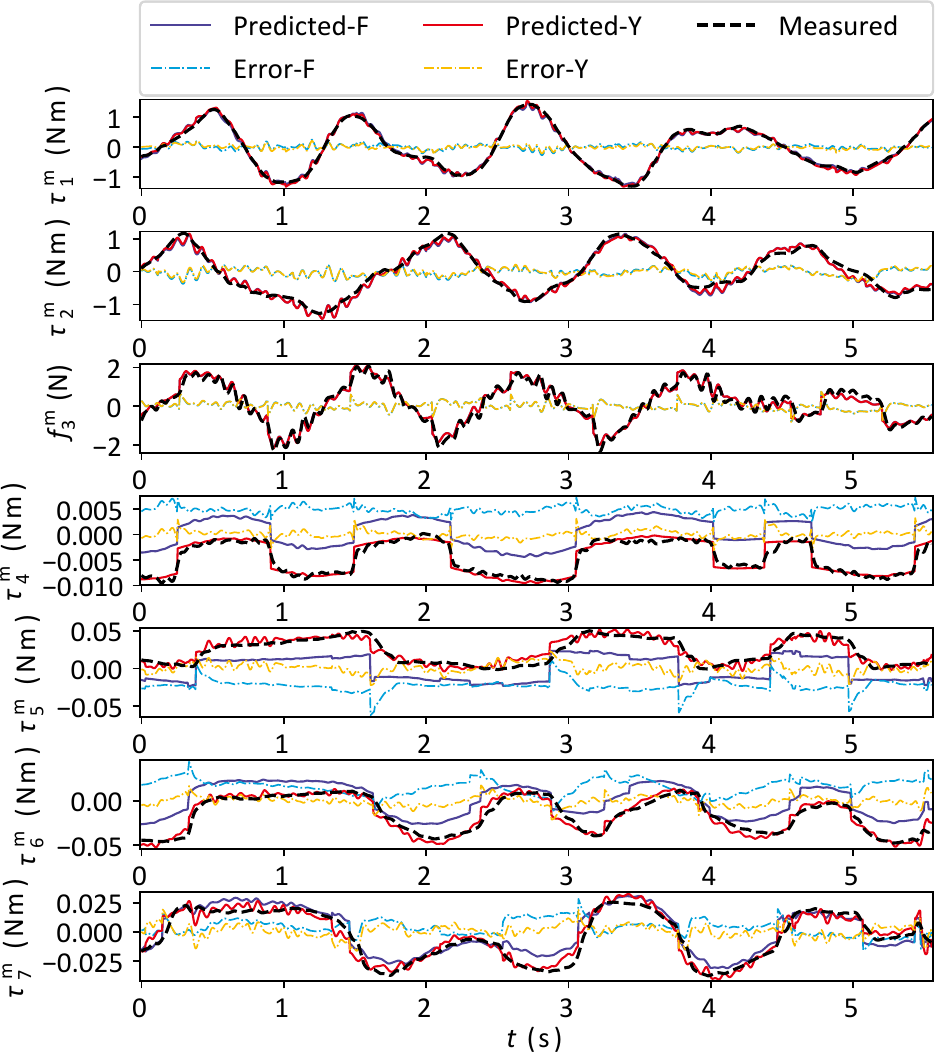}
\caption{\label{fig:psm_cvx_test} Measured and  predicted torques on the test trajectory of the PSM.}
\end{figure}
Fig. \ref{fig:mtm_cvx_test} and \ref{fig:psm_cvx_test} show the comparison of the measured and predicted torques on the test trajectories for the MTM and PSM, respectively. 
The relative prediction error of each motor joint is shown in Table \ref{table:relative_predict_error}.
The suffixes, -F and -Y, represent the modeling from \cite{fontanelli2017modelling} and our work, respectively.

For our proposed approach, the relative prediction errors of the first three motor joints of the MTM are less than $17.0\%$, which correspond to the Cartesian motion and most of the link inertia of the MTM. 
The large backlash from gearboxes and small link inertia of the last four joints make it hard to identify their dynamic parameters accurately. Hence, the relative prediction errors of the last four motor joints are relatively higher. 
Compared to the method from \cite{fontanelli2017modelling}, our proposed approach achieves  better overall identification performance. Particularly, incorporating the modeling of the nonlinear friction on joint 4 and the spring on joint 5 improves the identification performance for  joints 4 and 5, significantly. 

For our proposed approach, the relative prediction error of the first three motor joints of the PSM is less than $19.1\%$, which correspond to the Cartesian motion and most of the link inertia of the arm. 
The relative prediction errors of the last four motor joints are relatively larger since they are only modeled with motor inertia and frictions, and the magnitudes of the joint torques are very small. 
Compared to the method from \cite{fontanelli2017modelling}, our proposed approach achieves  similar identification performance for the first three joints while much better performance for the last four joints. 
This improvement is owed to the modeling of friction offset (see joints 4 and 5 in Fig. \ref{fig:psm_cvx_test}) and motor inertia.


\subsubsection{Identification with a Weight on the PSM}
The same identification procedure was performed with a standard 200 g weight (totally 205 g, with 5 g tapes added) firmly taped  on the top of the parallelogram of the PSM, i.e., link $2''$ (see Fig. \ref{fig:psm_geo}).
We listed all the seven  base parameters related to $m_{2''}$ in Table \ref{table:base_parameters}.
Since each complete symbolic base parameter is too long to show here, we only show part of it to illustrate the relation between the parameter and $m_{2''}$.

With the values of one parameter identified with and without the weight (i.e., $\hat{\delta}_b$ and $\hat{\delta}_b^w$), we estimated the mass of the weight as, $\hat{m}_{w} = (\hat{\delta}_b^w - \hat{\delta}_b)/c_{m_{2''}}$, where $c_{m_{2''}}$ is the coefficient of the corresponding $m_{2''}$ term. 
The relative estimation error of the weight was calculated by $\epsilon_w = |\hat{m}_{w}-205|/205$.
As shown in Table \ref{table:base_parameters}, the low $\epsilon_w$ was achieved through most parameters, except the $5^{\mathrm{th}}$ one whose $\epsilon_w$ is as high as 82.5\%.
This can be caused by identification noise.
The $c_{m_{2''}}$ of this parameter is only 0.00576, which is much smaller than the $c_{m_{2''}}$ of other parameters, and thus $\hat{m}_w$ is more sensitive to noise for this parameter. 
In summary, the overall accurate estimation of the mass of the weight further demonstrates the robustness of the proposed approach and package. 

\begin{table}[!tb]
    \caption{Comparison of the $m_{2''}$-Related Base Parameters Identified with and without the Weight}
    \label{table:base_parameters}
    \begin{center}
        \begin{tabular}{@{}l@{\hskip6pt}l@{\hskip6pt}l@{\hskip6pt}l@{\hskip6pt}l@{}}
        \toprule
        \textbf{base parameter related to $m_{2''}$} & $\hat{\delta}_b$  & $\hat{\delta}_{b}^w$ & $\hat{m}_{w}$ (g) & $\epsilon_w$ (\%)\\ \midrule
        $- 0.5 l_{2y}  - 0.072 m_{2''} + \hdots$                        &  0.06147         &   0.04668    &     204.6   & 0.2     \\
        $0.5 l_{2x} + 0.020 m_{2''} + \hdots$                      &  -0.01895        &   -0.01437    &     228.2     & 1.1   \\
        $0.5 l_{2y} + 0.072 m_{2''} + \hdots$                     &    -0.1324       &    -0.1177   &     203.2      & 0.9  \\
        $- 0.5 l_{2x} - 0.020 m_{2''} + \hdots$                       &  0.02080         &  0.01725     &  176.9     & 13.7      \\
        $L_{2xy} + 0.00576 m_{2''} + \hdots$                      &  -0.05245         &  -0.05038     &   35.9       & 82.5   \\ 
        $L_{2xx} - 0.0415 m_{2''} + \hdots$                      &   0.2068           &  0.1995      &   176.2 & 14.0\\
        $L_{2zz}  - 0.0415 m_{2''} + \hdots$                     & 0.2390             & 0.2307     & 198.0 & 3.4\\
        
        \bottomrule
        \end{tabular}
        \end{center}
        \end{table}
        

\section{Conclusion}\label{sec:conclusion}
In this work, an open-source software package for the dynamic model identification of the dVRK is presented\footnote{\url{https://github.com/WPI-AIM/dvrk_dynamics_identification}}. 
Link inertia, joint friction, springs, tendon couplings, cable force, and closed-chains are incorporated in the modeling. 
Fourier series-based trajectories are used to excite the dynamics of the dVRK, with the condition number of the regression matrix minimized. 
A convex optimization-based method is used to obtain  dynamic parameters subject to physical consistency constraints. 
Experimental results show the improvement of the proposed modeling and the robustness of the package.
Although this software package is developed for the dVRK, it is feasible to use it on other robots.

Despite the improvement of identification performance in our modeling compared to \cite{fontanelli2017modelling}, we can still observe substantial deviations between the measured and predicted torques. 
Although the convex optimization-based framework ensures the global optimality of identification results, it relies on the linearity of dynamic parameters w.r.t. joint torques \cite{sousa2014physical}.
As a result, nonlinear friction models considering presliding hysteresis, such as the Dahl model \cite{dahl1968solid}, which can potentially improve the modeling of electrical cables and tendon-sheath transmission, cannot be used in this package.
Moreover,  the present identification approach requires the computation of acceleration, which provides more information, however, requires correct handling of data filtering, compared to energy model-based methods \cite{gautier1988identification}.





%


\printbibliography 
%








\end{document}